%% file: main.tex
\documentclass[10pt,twocolumn,letterpaper]{article}

\usepackage{iccv}              
\usepackage[pagebackref,breaklinks,colorlinks,allcolors=iccvblue]{hyperref}

\usepackage{amsfonts}
\usepackage{times}
\usepackage{soul}
\usepackage{url}
\usepackage[utf8]{inputenc}
\usepackage[font=small]{caption}
\usepackage{graphicx}
\usepackage{amsmath}
\usepackage{amsthm}
\usepackage{booktabs}
\usepackage{algorithm}
\usepackage[switch]{lineno}
\usepackage{adjustbox}
\usepackage{algpseudocode}
\usepackage{newpxtext}
\usepackage{array}
\usepackage{multirow}
\usepackage{colortbl}

\usepackage{placeins}

\definecolor{iccvblue}{rgb}{0.21,0.49,0.74}

\def\paperID{14104} 
\def\confName{ICCV}
\def\confYear{2025}

\title{Federated Domain Generalization with Domain-specific Soft Prompts Generation}

\author{Jianhan Wu \thanks{Equal contribution}\quad Xiaoyang Qu \footnotemark[1] \quad Zhangcheng Huang\quad Jianzong Wang \thanks{Corresponding author is Jianzong Wang. This work was supported by the National Key Research and Development Program of China (Youth Scientist Project) under Grant No. 2024YFB4504300 and the Shenzhen-Hong Kong Joint Funding Project (Category A) under Grant No. SGDX20240115103359001.} \\
Ping An Technology (Shenzhen) Co., Ltd.\\
	{\tt\small wujianhan@mail.ustc.edu.cn, quxiaoy@gmail.com}\\
	{\tt\small william.huang@connect.polyu.hk, jzwang@188.com}
}

\begin{document}

\maketitle
\input{sec/0_abstract}    
\input{sec/1_intro}
\input{sec/2_related_work}
\input{sec/3_methodology}
\input{sec/4_experiment_setup}

\input{sec/5_experiments_results}
\input{sec/6_conclusion}
{
    \newpage
    \small
    \bibliographystyle{ieeenat_fullname}
    \bibliography{main}
}

\end{document}

%% file: sec/0_abstract.tex
\begin{abstract}
Prompt learning has become an efficient paradigm for adapting CLIP to downstream tasks. Compared with traditional fine-tuning, prompt learning optimizes a few parameters yet yields highly competitive results, especially appealing in federated learning for computational efficiency. engendering domain shift among clients and posing a formidable challenge for downstream-task adaptation. Existing federated domain generalization (FDG) methods based on prompt learning typically learn soft prompts from training samples, replacing manually designed prompts to enhance the generalization ability of federated models. However, these learned prompts exhibit limited diversity and tend to ignore information from unknown domains. We propose a novel and effective method from a generative perspective for handling FDG tasks, namely federated domain generalization with domain-specific soft prompts generation (FedDSPG). Specifically, during training, we introduce domain-specific soft prompts (DSPs) for each domain and integrate content and domain knowledge into the generative model among clients. In the inference phase, the generator is utilized to obtain DSPs for unseen target domains, thus guiding downstream tasks in unknown domains. Comprehensive evaluations across several public datasets confirm that our method outperforms existing strong baselines in FDG, achieving state-of-the-art results.
\end{abstract}

%% file: sec/1_intro.tex
\section{Introduction}
Pre-trained vision-language models (VLMs) \cite{jia2021scaling}, such as Contrastive Language Image Pretraining (CLIP) \cite{radford2021learning,lu2023fedclip,hu2024reclip}, have emerged as a powerful technique to adapt to downstream tasks. By introducing learnable prompts \cite{Bai2024DiPrompTDP,zhao2023fedprompt,guo2023promptfl} into the text or vision encoders of CLIP, this approach allows the model to dynamically adjust its representations with minimal fine-tuning. Unlike traditional fine-tuning methods \cite{zaken2021bitfit} that require extensive parameter updates, prompt learning optimizes only a small fraction of parameters, significantly reducing computational costs while maintaining high performance. This approach has been extensively adopted in federated learning (FL) \cite{mcmahan2017communication,mohri2019agnostic} paradigms, particularly in scenarios characterized by substantial computational and communication overheads between server and client devices. While federated learning based on prompt learning demonstrates substantial privacy and computational benefits, a key challenge persists: data on each node is often collected in a non-IID manner in real-world scenarios, leading to domain shift \cite{liu2023co,tan2024heterogeneity} between nodes. For example, one device may take photos mostly indoors, while another mostly outdoors \cite{peng2019federated}.	The presence of the domain shift issue renders the direct application of learning with hand-crafted prompts (such as "a photo of a [class]") in federated learning incapable of achieving satisfactory outcomes \cite{wei2024learning}.

To address the domain shift problem in FL, federated domain generalization (FDG) \cite{song2020privacy,liu2023co,fang2024source,liu2024ufda} further considers that the test data (target domain) is unavailable during decentralized training from distributed source domains. Some methods have optimized soft prompts \cite{qiu2023text,jia2022visual} across different clients by treating them as learnable vectors and refining them via backpropagation. For example, Promptfl \cite{guo2023promptfl} shares prompt vectors instead of model parameters to enhance communication efficiency and privacy in FL. pFedPG \cite{yang2023efficient} introduced a domain bank mechanism to embed textual domain information into soft prompts. Recently, ADAPT \cite{wei2024learning} and FedTPG \cite{qiu2024federated} used lightweight neural networks (e.g., adapter \cite{hu2021lora}) to optimize soft prompts conditioned on images through residual or concatenation mechanisms \cite{cui2024harmonizing}, which improves generalization in a scalable manner. Despite recent advances, these methods either rely heavily on the diversity of client-specific data or use a simple adapter layer to estimate directly the relationship between soft prompts and images, leading to the learned prompts failing to fully capture the complex and diverse cross-domain feature \cite{bai2025soft}.

To tackle the aforementioned challenges, we propose a novel and effective federated domain generalization method with domain-specific soft prompt generation (FedDSPG). Our approach reconceptualizes the prompt learning mechanism from a generative perspective, leveraging a generative model to produce domain-specific soft prompts (DSPs).
Through an adversarial game manner between the generator and the discriminator, the generated DSPs exhibit greater diversity and generalization in both form and semantics. This enables our prompt generator model to acquire more generalized and robust features, facilitating adaptation to complex downstream tasks. The architectural details of our proposed approach are comprehensively depicted in Figure \ref{fig: whole structure}, comprising a training phase with two steps and an inference phase. Specifically, in the first step of the training phase, we avoid using hand-crafted prompts or soft prompts that lack sufficient attribute focus and specific content design. We elaborately design DSPs, composed of domain and class name vectors that map domain and content information, representing the optimal prompts for images from each client. In the second step, images and their corresponding DSPs are fed into a simple yet effective conditional generative model \cite{ding2021ccgan}, which is trained to further optimize the DSPs by learning domain and content knowledge of images. Given the significant variance in tunable prompt embedding parameters across data-heterogeneous clients, the traditional FedAvg \cite{mcmahan2017communication} aggregation method yields suboptimal results (see Table \ref{tab:abl-tpu} for comparative experiments). To mitigate this, we propose using a momentum-based method to aggregate tunable parameters in the first training step. During the inference stage, the generative model's generator component will generate DSPs for target domain images, which subsequently facilitate predictions in downstream tasks.

Our key contributions are summarized below:
\begin{itemize}
\item We are the first to optimize prompt learning from a generative perspective in FL and propose a novel soft prompt generation paradigm for FDG tasks.
\item Unlike the current CLIP-based methods that rely on hand-crafted or naïve soft prompts, we devise domain-specific soft prompt structures that significantly enhance generalization to unseen domains. Furthermore, we introduce a momentum-based aggregation mechanism to consistently refine performance.
\item Comprehensive empirical evaluations on standard benchmark datasets corroborate that our proposed method achieves state-of-the-art performance. Additionally, comprehensive ablation research indicates the role played by each strategy in the increased efficacy.
\end{itemize}

%% file: sec/2_related_work.tex
\section{Related Work}
\begin{figure*}
\centering
\includegraphics[width=1.0\linewidth]{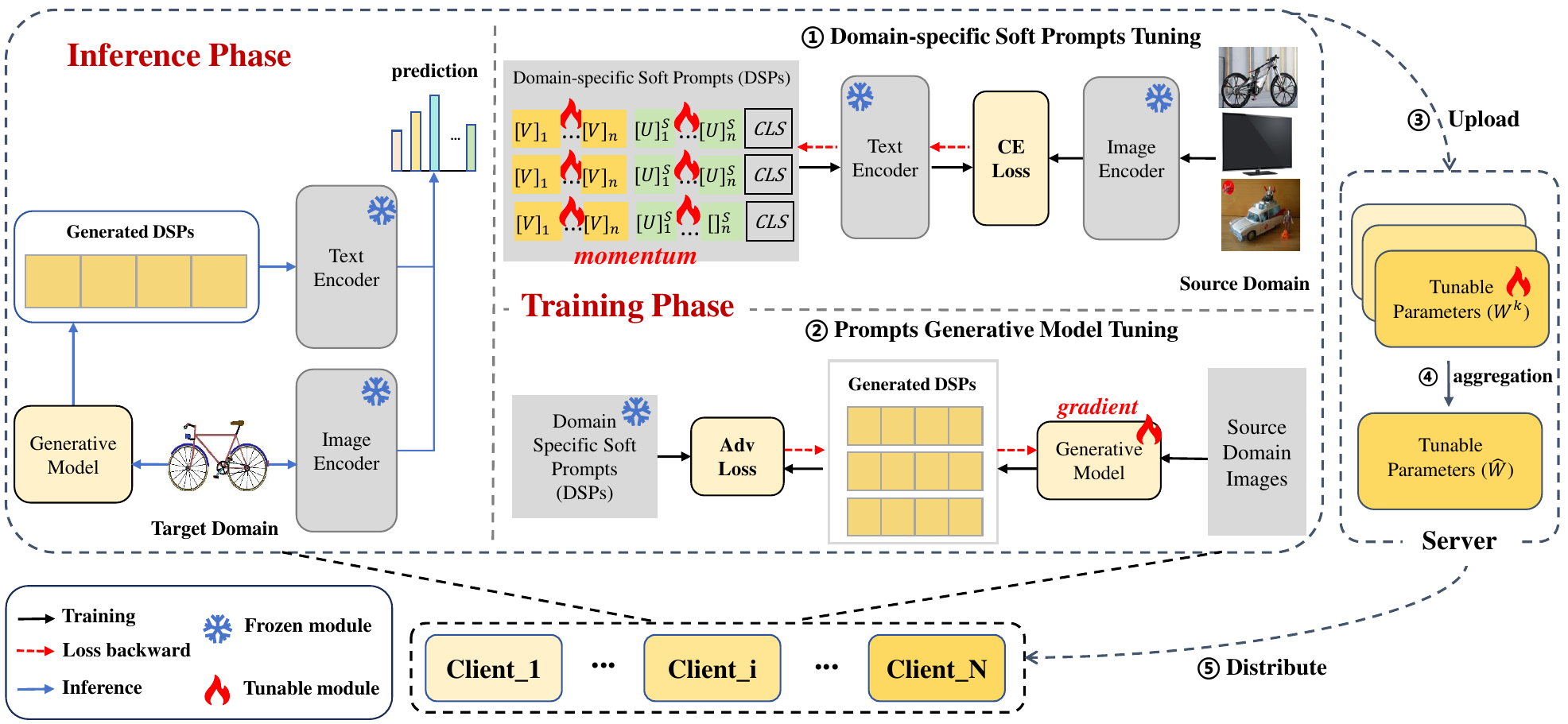}
\caption{The overall training structure of FedDSPG consists of two main stages: domain-specific prompts tuning and domain prompt generation. The first stage produces DSPs with compositional soft prompting technique \cite{nayak2022learning}, which encode the content and style information with large-scale pretrained VLMs. In the second stage, a GAN-based training method is employed to learn the latent data distribution, thereby generating soft prompts that adhere to the intrinsic data patterns. For the inference stage, the generative model's generator component will generate DSPs for target domain images, producing enhanced predictions for downstream tasks. Additionally, the momentum-based aggregation is conducted to enhance model generalization for FDG.}
\label{fig: whole structure}
\vspace{-0.2cm}
\end{figure*}

\subsection{Prompt Learning for Vision Language Models}
Prompt learning, also known as prompt tuning, provides an efficient alternative to conventional fine-tuning approaches for pre-trained models. By focusing on tuning only learnable prompt tokens attached to the input of models like CLIP. CoOp \cite{zhou2022learning} first introduced soft prompts into vision-language models, demonstrating that appropriate prompts can enhance performance in downstream tasks. CoCoOp \cite{zhou2022conditional} addresses the overfitting issue in CoOp through conditioned prompt learning, specifically residual prompts. DPL \cite{zhang2023domain} achieved effective domain generalization by training a lightweight prompt generator (a linear network). FUZZLE \cite{Shi2024UnsupervisedDA} integrates fuzzy C-means clustering into prompt learning to optimize cross-domain alignment between source and target distributions. In these studies, soft prompts are typically generated by simple tunable models (e.g., adapter) that predominantly rely on the diversity of client-specific data, which yields unsatisfactory results for the unseen domain. The research most pertinent to ours is SPG \cite{bai2025soft}, which employs conditional adversarial generative networks to generate prompts but lacks domain-specific design, resulting in suboptimal performance in federated learning scenarios. We not only meticulously designed DSPs but also tailored the federated aggregation algorithm to better execute FDG.
\subsection{Federated Domain Generalization}
FDG aims to address domain shifts between source and target domains in federated settings where data is distributed across clients. Traditional FDG methods generally fall into two types. The first is domain alignment approaches that focus on learning domain-invariant representations by reducing source domain discrepancies. For instance, in KD3A \cite{feng2021kd3a}, domain differences are reduced by explicitly matching the mean embeddings of data from different domains. These methods' disadvantage lies in stringent assumptions about data distribution that are hard to satisfy in practice. The second type involves data augmentation methods that aim to expand the training data distribution to enhance out-of-distribution generalization. Additional strategies include meta-learning \cite{alsulaimawi2024meta}, adversarial learning \cite{zhao2023federated}, and more. Recently, PromptFL \cite{guo2023promptfl}, ADAPT \cite{wei2024learning}, and FedTPG \cite{qiu2024federated} introduced prompt learning to extract semantic information, aligning domain information on implicit textual features to improve the model's generalization capacity. To further advance this field, we conduct an in-depth exploration of this efficient paradigm. Departing from the previous prompt learning designs that rely on hand-crafted prompts or simple neural network, we present a novel and effective FDG structure that leverages generative modeling techniques to learn domain-specific soft prompts, enhancing the generalization of prompts to unseen target domains.

%% file: sec/3_methodology.tex
\section{Methodology}
In this section, we first introduce the preliminary knowledge of problem setup of FDG. Following that, the proposed method FedDSPG will be introduced in detail, which includes two training steps of domain-specific soft prompts tuning and prompts generative model tuning, as well as federated parameter aggregation and federated inference.

\subsection{Preliminaries}
In the standard FDG problem, there is a source domain dataset $ \mathcal{D}_S = {(\textbf{x}^{s}_i, \textbf{y}^{s}_i)}^n_{i=1} $ with labeled data, where $(\textbf{x}^{s}_i, \textbf{y}^{s}_i)$ is the data distribution for client $i's$ input space and corresponding labels, with n being the number of source domains. The target domain dataset can be defined as $ \mathcal{D}_T = {(\textbf{x}^{t}_i, \textbf{y}^{t}_i)}^m_{i=1} $, where the number of target domains $m$ is typically configured as one per client for experiment evaluation. FDG seeks to develop models that can achieve robust performance on target domains while being trained exclusively on distributed source domain data. The primary challenge lies in addressing the domain shift between source and target domains, which is exacerbated by the non-IID nature of federated data across clients.

 Taking image classification as an example, let $X$ and $Y$ represent the input and label spaces, respectively. Previous domain generalization methods concentrated on training a domain-invariant model for the mapping $F:X \xrightarrow{}Y$ from images to labels. Prompt learning methods embed learnable prompt vectors $V$ into the input to leverage the generalization ability of VLMs, reformulating the mapping as $F:(X, V)\xrightarrow{}Y$. Unlike existing prompt learning methods that focus primarily on fine-tuning the prompt vectors $V$, we propose a novel prompt learning paradigm that utilizes a generative model $G$ to generate soft prompts, reformulating the mapping as $F:(X, G(X))\xrightarrow[]{}Y$. The objective is to develop a generative framework that simultaneously captures both domain-specific and domain-invariant representations, enabling the dynamic generation of generalized prompts for unseen target domains.

\begin{figure}
\centering
\includegraphics[width=1.0\linewidth]{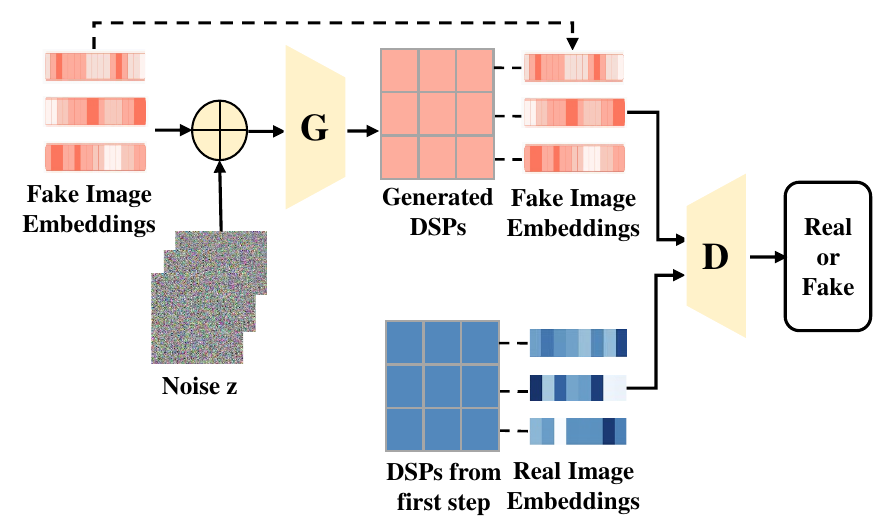}
\caption{The structure design of the second training step is based on CcGAN. In the generator, input noise $z$ is concatenated with fake image embeddings to form the joint latent representation, which is used to generate DSPs. The discriminator distinguishes the authenticity of the generated DSPs from the real image.}
\label{fig: gan structure}
\vspace{-0.2cm}
\end{figure}

\subsection{Two Training Steps}
FedDSPG is implemented in two steps: domain-specific soft prompts tuning and prompts generative model tuning. Domain-specific prompts tuning aims to embed domain and class information into prompts, and the goal of prompts generative model tuning is to incorporate such information into the generative model. Notably, domain distribution is learned by the generative model, enabling the generation of instance-specific prompts for each image, guaranteeing prompt diversity and integrating domain-specific information. Below are the detailed designs of FedDSPG.

\textbf{Domain-specific Soft Prompts Tuning.} 
We adopt CLIP \cite{radford2021learning} as our backbone, leveraging its contrastive learning framework trained on extensive image-text pairs to establish a joint embedding space through dedicated image ($f$) and text ($g$) encoders. By describing the input text as structured prompt $t_i$: "a photo of a [CLASS]", the similarity of image representations in the same category $i$ can be maximized to align semantic representations:
\begin{equation}
P(\hat{y} = i | x) = \frac{\exp(Sim(g(t_i), f(x) )/ \tau)}{\sum_{k=1}^{K} \exp(Sim(g(t_k), f(x)) / \tau)}
\end{equation}	

Where $Sim(\cdot)$ is the cosine similarity and $\tau$ denotes temperature parameter. Since the soft prompts of previous work (e.g., CoOp \cite{zhou2022learning} and SPG \cite{bai2025soft} $ t_i = [v]_1 [v]_2 \dots [v]_{M_1} [\text{CLASS}]_i$) cannot deal well with the distribution shift between domains, we design domain-specific soft prompts to serve as the optimal prompt representations for images from each client. Specifically, each image from a client source domain is associated with unique DSPs (seen in Figure \ref{fig: whole structure}). The tunable vectors of DSPs contain two parts, a domain-invariant context and a domain-specific context, denoted as:
\begin{equation}
t^d_{i} = [v]_1 [v]_2 \dots [v]_{M_1} [u]^d_{1} [u]^d_{2} \dots [u]^d_{M_2} [\text{CLS}]_i
\end{equation}

Where $ [u]^d_{M_2}$ denotes domain-specific tokens, which have the same dimension as word embeddings. $M_1$ and $M_2$ are the number of context tokens and domain tokens incorporated in the prompt structure, respectively. These DSPs are optimized through cross-entropy loss minimization during training on domain-specific data:
\begin{equation}\begin{aligned}\label{func:dpl}
{\mathbf{v}_{c_j}^{d_i}}^*=\arg \min _{\mathbf{v}} \mathbb{E}_{\mathbf{x}^{d_i}_j, {y}^{d_i}_j}\left[-\log p\left({y}_j^{d_i} \mid \mathbf{x}^{d_i}_j, \mathbf{t}_{j}^{d_i}\right)\right]
\end{aligned}\end{equation}

Where $(\mathbf{x}^{d_i}_j$, ${y}^{d_i}_j)$ represents the image-label pairs from client $j's$ training data in $i$-th domain. DSPs integrate domain and class information in a more coherent and structured manner than prior approaches. These DSPs constitute optimized prompt representations for each client domain, encoding comprehensive domain characteristics, which encapsulate rich domain and class information for training a generative model in the second training step.

\textbf{Prompts Generative Model Tuning.}
To learn the domain and class knowledge in DSPs, considering the communication cost requirements (e.g., methods based on diffusion models \cite{ho2020denoising} require multi-step denoising, resulting in high communication costs. The corresponding experiments can be found in the supplementary materials.) and input data continuity of federated learning, we employ an efficient generative model—Continuous Conditional Generative Adversarial Network (CcGAN) \cite{ding2021ccgan} to validate the effectiveness of our approach. Similar to conditional GAN \cite{mirza2014conditional}, we manipulate the generation process by constraining the additional information (i.e., DSPs generated in the first training step) of the CcGAN. The goal is to learn the generator's distribution $p_g$ over DSPs $\mathbf{v}_{c_j}^{d_i}$, effectively transferring generalization capabilities from prompts to the generator itself. The architecture consists of two core components: a generator $G$ that synthesizes outputs conditioned on auxiliary information and a discriminator $D$ that assesses the authenticity of real data and fake outputs. These components facilitate an adversarial process that enhances the model's generative performance.

Specifically, the detailed training process of the prompts generative model is shown in Figure \ref{fig: gan structure}. The generator concatenates noise variables $\textbf{z}$ with fake-batch embeddings $f(\textbf{x})$ to form joint latent codes $[\textbf{z}, f(\textbf{x})]$, where $\textbf{x}$ is randomly sampled from client data. Then the generator subsequently produces the corresponding DSPs by transforming the joint latent representation $[\textbf{z}, f(\textbf{x})]$. In the discriminator, we input both the genuine image embeddings $f(\textbf{x})$ from real batches and the generated DSPs with their associated image embeddings $f(\textbf{x})$ to distinguish real from fake. Therefore, the two-player min-max game objective function $V(G, D)$ for CcGAN can be formally expressed as:

\begin{equation}\begin{aligned}\label{func:adv_loss}
\min _G \max _D V(G, D)
&=\mathbb{E}_{\mathbf{v} \sim p_{\text {v}}(\mathbf{v})}[\log D(\mathbf{v} \mid f(\mathbf{x}))] \\
&\quad + \mathbb{E}_{{\mathbf{z} \sim p_{\text {z}}}(\mathbf{z})}[\log (1-D(G(\mathbf{z} \mid f(\mathbf{x})) ))]
\end{aligned}
\end{equation}

The process of prompts generative model tuning is to learn coarse-to-fine DSPs through an adversarial manner. This design not only promotes diversity in prompt generation but also significantly improves generalization across tasks and domains, facilitating robust performance in both seen and unseen conditions.

\subsection{Federated Parameters Aggregation} \label{aggregation}
We employ distinct aggregation strategies for the learnable parameters of prompt tuning and the generative model. Assuming a centralized parameter server for uniform aggregation—though our method also supports decentralized communication—clients upload their respective soft text embedding parameters and generative model parameters to the server during each communication round. The server averages the parameters from all n clients and redistributes them. While this parameter aggregation paradigm appears viable for our method, it introduces a minor issue for soft text embeddings. Specifically, given that FDG addresses domain shifts across clients, the domain components of DSPs vary significantly between clients. We observe that direct averaging causes abrupt changes in soft text embedding parameters, negatively impacting model performance. To mitigate this, we introduce momentum updates for soft prompt parameters, utilizing exponential moving averages to maintain parameter continuity and prevent destabilizing abrupt changes during the optimization process. The aggregation formula is as follows:
\begin{equation}
    [w]^k = \alpha [w]^{k-1} + (1-\alpha)[w]^{k-2}
\end{equation}

Where $[w]^k$, $[w]^{k-1}$ and $[w]^{k-2}$ denote the tunable soft prompt embedding parameters at the $k$, $k-1$ and $k-2$ step, and $\alpha \in [0, 1]$ serves as a hyperparameter governing the smoothing intensity of DSPs. Comprehensive ablation studies examining the impact of momentum updates are presented in table \ref{tab:abl-tpu}.

\subsection{Inference with Generator}
The generator trained via CcGAN is utilized to generate DSPs for inference on each target image. The predicted probability that a target domain image $x$ belongs to the $i$-th class is formally expressed as:

\begin{equation}\begin{aligned}\label{func:pro}
p(y=i \mid \mathbf{x})=\frac
{\exp (\langle w_i, f(\mathbf{x}) \rangle / \tau)}
{\sum^K_{j=1} \exp (\langle w_j, f(\mathbf{x}) \rangle / \tau)}, 
\end{aligned}\end{equation}

\begin{equation}\begin{aligned}\label{func:w}
\text{s.t.} ~~~~~~~~ w_j = g([G(\mathbf{z} \mid f(\mathbf{x})), c_j])
\end{aligned}\end{equation}

Where $\tau$ represents the temperature parameter that modulates the sharpness of the output distribution, $g$ and $f$ represent the text and the image encoder respectively, and $K$ denotes the total number of classes. The term $G(\mathbf{z} \mid f(\mathbf{x}))$ signifies the generated DSPs by the generator. We use $c_j$ to represent the $j$-th tokenized class label, with $[\cdot,\cdot]$ denoting the standard concatenation operation. In the generative paradigm, our approach achieves enhanced adaptability and dynamic prompts generation, which are essential for accommodating diverse input distributions and downstream task requirements. 

%% file: sec/4_experiment_setup.tex
	\begin{table*}
	\begin{adjustbox}{width=\textwidth}
		\renewcommand\arraystretch{1.4}
		\begin{tabular}{ l | c  c  c  c  c | c  c  c  c  c  c  c | c c c c c}
			\toprule
		 	\textbf{Dataset} &\multicolumn{5}{c|}{\textbf{Office-Home}} & \multicolumn{7}{c|}{\textbf{DomainNet}}&\multicolumn{5}{c}{\textbf{PACS}}  \\
			\midrule
			\textbf{Target Domain}& Re & Pr & Cl & Ar & \textbf{Avg} &  Clp & Inf & Pnt & Qdr & Rel & Skt & \textbf{Avg} & P & A & C & S & \textbf{Avg} \\
			\midrule

			CoOp\cite{zhou2022learning} &   82.9   &   82.8   &    59.3   &   72.1   &   74.3   &   61.6   &   22.2   &   52.6   &   12.1   &   67.7   &   48.6   &   44.1  &94.1&91.5&95.6&84.9&91.5\\
			CoCoOp\cite{zhou2022conditional} &   85.0   &   81.1   &   62.5   &   68.6   &  74.3   &   72.4   &   23.3   &   60.8   &   16.4   &   72.7   &   60.5   &   51.0  &97.5&94.6&97.9&87.5&94.4 \\
             FedVPT-D\cite{jia2022visual}&    85.0   &  85.2   &   61.1     &  76.0  & 76.8  & 63.3   &  43.0     &   74.8    &  54.8    &  87.2    &   67.1  & 65.0&99.2&96.5&98.1&87.2& 95.3 \\
             FedCLIP\cite{lu2023fedclip}&    87.7   &  87.5   &   63.6     &  78.0  & 79.2  & 65.1  &  46.5   &  75.9   & 56.1   &  90.2  & 67.9  &67.0&99.8&96.3&97.9&85.6&94.9  \\
              PromptFL\cite{guo2023promptfl}&    89.6  &  88.2   &   71.6    &  79.8 & 82.3  & 72.8  &  50.3  &   84.2 &  60.4  &  92.1  &  67.7  & 71.3 &\textbf{99.9}&97.1&99.0&90.6&96.7\\

              ADAPT\cite{wei2024learning} &   90.3   &  90.5  &   68.2      &  82.6 & 82.9  &  \textbf{77.5}   &  63.1    &   70.5   &  41.6    &   85.7  &  72.1    & 68.4&\textbf{99.9}&98.0&99.1&91.7&97.2\\
              FedTPG\cite{qiu2024federated}&   89.0  &   89.7  &   71.1 &  84.7 & 83.6 & 74.0  & 52.1   &   85.1  &  62.1  &  94.0  & 68.6  &     72.7&99.8&97.8&99.2&90.8&96.9 \\
                \midrule
			\textbf{FedDSPG} &   \textbf{90.5}   &   \textbf{91.0}   &  \textbf{ 73.8}   &   \textbf{85.8}   &   \textbf{85.3}   &  74.7 &   \textbf{63.5}   &   \textbf{85.5}   &   \textbf{63.0}   &   \textbf{94.2}   &  \textbf{72.6}  &   \textbf{75.6} &\textbf{99.9}&\textbf{98.1}&\textbf{99.3}&\textbf{92.2}&\textbf{97.4} \\
			\bottomrule      
		\end{tabular}
	\end{adjustbox}
	\caption{Accuracies (\%) for the image classification task on Office-Home, DomainNet, and PACS datasets using CLIP with ViT-B/16 image encoder as frozen backbone. The domains displayed here are the target domains, and the others are the source domains; the best result is indicated in bold.}
	\label{tab: OfficeHome and DomainNet}
	\vspace{-0.2cm}
\end{table*}
\section{Experimental Setup}
\textbf{Datasets.} We conducted extensive experiments for image classification on domain generalization benchmark datasets: (1) Office-Home, a large-scale visual cross-domain dataset with 65 classes across four visually distinct domains, i.e., Art (Ar), Clipart (Cl), Product (Pr), and Real World (Re) \cite{venkateswara2017deep}. (2) DomainNet, a large-scale domain generalization benchmark dataset with 345 classes distributed among six domains, i.e., Clipart (Clp), Infograph (Inf), Painting (Pnt), Quickdraw (Qdr), Real (Rel), and Sketch (Skt) \cite{peng2019moment}. (3) PACS, featuring approximately 10,000 images from seven categories across four domains, including Photo (P), Sketch (S), Cartoon (C), and Art (A) \cite{li2017deeper}. For rigorous evaluation, we adopt the leave-one-domain-out protocol, designating each domain sequentially as the target while utilizing the remaining domains as source data.

\textbf{Implementation details.} In FDG, all participating nodes maintain identical architectural configurations while ensuring complete domain data isolation between nodes. The data processing and division are followed \cite{ge2023domain} on the image generalization task. For our FedDSPG method, we employ a frozen CLIP backbone (ViT-B/16 \cite{alexey2021image}) pre-trained on large-scale vision-language data. The text and visual prompt tokens share a unified dimensionality of 512. For the learning strategies, in the first training step, the DSPs are trained by using the Adam optimizer with the learning rate of 1$\times 10^{-5}$, batch size of 32, and the context lengths $M_1$ and $M_2$ of the prompt number are 4. In the second training step, the CcGAN is trained by using the AdamW optimizer with weight decay 2$\times 10^{-5}$ and $\beta_1 = 0.9$, $\beta = 0.999$), starting with a learning rate of 1$\times 10^{-4}$ for the OfficeHome and DomainNet datasets. The 100 epochs are trained for both our method and the baseline models for accelerating convergence, then the FedAvg and momentum-based aggregation algorithms are executed after every epoch. 8 Tesla-V100 with 16GB memory are used to experiment. 

\textbf{Baseline models.} We compared three distinct SOTA prompt learning approaches: (1) methods focusing on designing trainable models with small-scale parameters on the client side, such as CoOp \cite{zhou2022learning}, CoCoOp \cite{zhou2022conditional}, FedVPT-D\cite{jia2022visual}, FedCLIP \cite{lu2023fedclip}, and FedTPG \cite{qiu2024federated}; (2) methods that design prompt generators on the server side to reduce communication costs, such as PromptFL \cite{guo2023promptfl}; and (3) methods that emphasize both semantic and domain-specific prompts, such as ADAPT \cite{wei2024learning}. To show fair competition, the results of subsequent experiments were either the result of their papers or reproduced from source code provided by their authors.

%% file: sec/5_experiments_results.tex
\section{Experiments Analysis}
\subsection{Federated Domain Generalization Performance}
FedDSPG achieved effective FDG in scenarios where target domain distributions diverge significantly from source domains. Quantitative results across three benchmark datasets are presented in Table \ref{tab: OfficeHome and DomainNet} following the "leave-one-out" protocol. In general, our method outperforms all baseline methods to varying degrees. We gleaned two primary observations from the results. Firstly, our approach outperforms all baselines, achieving SOTA performance, which demonstrates the effectiveness of our proposed approach in addressing federated domain generalization challenges. This establishes FedDSPG as a highly referenceable benchmark for future research on FDG. Secondly, existing methods, including methods focusing on designing small-scale parameters models (FedCLIP, FedTPG) and methods generating prompts on the server side (PromptFL), exhibit notable performance gaps. We suppose that the possible reason stems from two primary factors: (1) inadequate incorporation of domain-specific information in their prompt design strategies, and (2) insufficient adaptability of generated prompts to unseen target domain images. These potential risks are mitigated in our approach, as it designs domain-specific soft prompts structure and trains a specific generative model, enhancing the generalization and diversity of prompts to unseen domains.
\begin{figure*}
    \centering
    \includegraphics[width=1.0\linewidth]{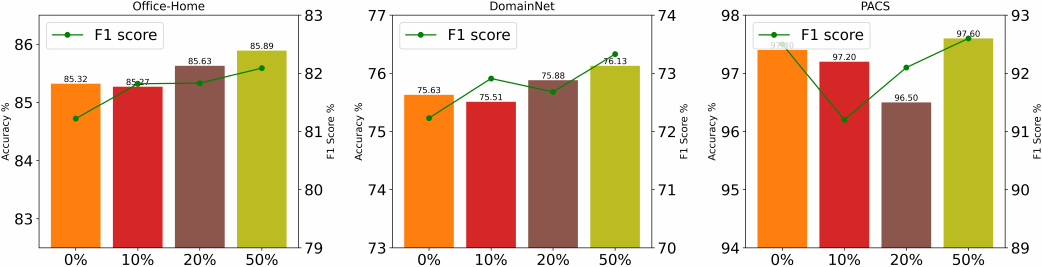}
    \caption{The accuracy of FedDSPG with different data distribution.}
    \label{fig: result of data distribution}
   \end{figure*} 
   
\subsection{Cross-Dataset Transfer}

  \begin{table*}
        \centering
        \renewcommand\arraystretch{0.95}
        \setlength\tabcolsep{12pt}  
        \resizebox{\textwidth}{!}{
        \begin{tabular}{@{}c|c|ccccc|ccccc@{}}
            \toprule
               &\textbf{Source}       & \multicolumn{10}{c}{\textbf{Target}}                      \\ 
               \midrule
            &\textbf{DomainNet} & Re & Pr & Cl & Ar  & \textbf{Avg}&P&A&C&S&\textbf{Avg}\\ \midrule
            
           FedTPG   & 72.7     &81.3      & 78.5      & 66.1      & 76.4      & 75.6&91.2&93.8&91.6&86.1&90.7  \\
                Ours   & 75.6     &86.0      & 82.5     & 70.4      & 80.4      & 79.8    &94.5&95.3&93.5&88.6&93.0\\
                \midrule
            \textbf{$\triangle$}   & \textbf{+3.9}      & \textbf{+4.7 }    & \textbf{+4.0}      & \textbf{+4.3  }    &\textbf{ +4.0}   &  \textbf{ +4.2}&\textbf{+3.3}&\textbf{+1.5}&\textbf{+1.9}&\textbf{+2.5}&\textbf{+2.3} \\ \bottomrule
        \end{tabular}
    }
        \caption{The results for cross-dataset transfer experiment. The model is trained on the DomainNet dataset and inference on the Office-Home and PACS dataset. }
        \vspace{-0.3cm}
    \label{tab: cross-dataset}
    \end{table*}

\begin{table*}
\renewcommand\arraystretch{1}
\centering
\resizebox{\textwidth}{!}{
\begin{tabular}{
>{\centering\arraybackslash}m{3cm}
>{\centering\arraybackslash}m{1cm}
>{\centering\arraybackslash}m{1.6cm}
>{\centering\arraybackslash}m{2.6cm}
>{\centering\arraybackslash}m{1.6cm}
>{\centering\arraybackslash}m{1cm}
>{\centering\arraybackslash}m{1.2cm}
>{\centering\arraybackslash}m{1cm}
>{\centering\arraybackslash}m{1.0cm}
>{\centering\arraybackslash}m{1.3cm}}
\toprule
\multirow{2}{*}{Method} 
    & \multicolumn{4}{c}{Prompt Type} &  &  &  &  &  \\
    \cmidrule(r){2-5}
    & Fixed & Learnable & Domain-specific & Generative & Re & Pr & Cl & Ar & \textbf{Avg} \\
\midrule

 PromptFL \cite{guo2023promptfl}& $\checkmark$  & $\checkmark$  &  &  & 89.6 & 88.2 & 71.6 & 79.8 & 82.3  \\
 ADAPT \cite{wei2024learning} &  & $\checkmark$  & $\checkmark$ &   & 90.3 & 90.5 & 68.2 & 82.6 & 82.9  \\
 FedTPG \cite{qiu2024federated}&   & $\checkmark$ &  &  & 89.0 & 89.7 & 71.1 & 84.7 & 83.6 \\
 Ours-HDP & $\checkmark$  &   & & $\checkmark$ & 86.1 & 81.4 & 70.2 & 73.2 & 77.7 \\
Ours-CSP&  & $\checkmark$  &   & $\checkmark$ & 90.9 & 90.8 & 73.1 & 84.1 & 84.6 \\
Ours-WGM &  & $\checkmark$  & $\checkmark$  &  & 88.1 & 87.8 & 71.6 & 82.4 & 82.5 \\
Ours &  & $\checkmark$  &  $\checkmark$   & $\checkmark$  & \textbf{90.5} & \textbf{91.0} & \textbf{73.8} & \textbf{85.8} & \textbf{85.3} \\
\bottomrule
\end{tabular}
}
\caption{Comparisons with different design of prompt on the Office-Home dataset for FDG performance.}
\label{abla_prompt1}
 \vspace{-0.3cm}
\end{table*}
Having demonstrated FedDSPG's capability for domain generalization within a single dataset, we further conducted experiments on cross-dataset domain generalization. This represents a more challenging task, as the fundamental characteristics can exhibit significant variations across different datasets. Specifically, we adopt a rigorous experimental protocol where DomainNet serves as the source domain, while Office-Home and PACS function as target domains. We compared FedDSPG with FedTPG (best-performing baseline), with the primary results detailed in Table \ref{tab: cross-dataset}. Overall, both methods experienced a decline in performance when confronted with the gap between datasets. However, FedDSPG consistently outperformed FedTPG on the target dataset by a substantial margin. This further underscores our method's remarkable capability to enhance model generalization across diverse domains.

\subsection{Data Distribution Analysis}
Considering client data distributions are inherently non-IID in federated learning scenarios, we conducted a comprehensive empirical analysis examining how varying data distributions affect FedDSPG's performance. Specifically, we varied the overlap ratio r of data domains across clients, setting it to 0\%, 10\%, 20\%, and 50\%. Specifically, an overlap ratio of $ r\%$ implies that $ r\%$ of the domains are shared among multiple clients, while the remaining $1- r\%$ of the domains are unique to individual clients.
Figure \ref{fig: result of data distribution} shows F1/accuracy across different domain distribution settings. The model exhibits greater stability on the Office-Home dataset compared to the other evaluation datasets, and a slight performance improvement is observed only when the overlap ratio reaches 50\%. We attribute this enhanced stability to the higher number of domains, which likely endows the model with greater robustness against distribution shifts. This finding underscores the importance of domain diversity in enhancing the generalization and stability of federated learning models.

\subsection{Ablation Study}
\begin{table*}[ht]
\centering
\renewcommand\arraystretch{1.0}
\setlength\tabcolsep{7pt}  
\begin{tabular}{cccccccccccc}
\toprule
\textbf{$\alpha$} & 0.0 & 0.1 & 0.2 & 0.3 & 0.4 & 0.5 & 0.6 & 0.7 & 0.8 & 0.9 & 1.0 \\
\midrule
\textbf{Accuracy (\%)} & 84.15 & 84.89 &\textbf{ 85.30 }& 85.28 & 84.56 & 84.32 & 84.67 & 84.01 & 83.95 & 83.50 & 83.39 \\
\bottomrule
\end{tabular}
\caption{Accuracy of FedDSPG with different coefficient $\alpha$.}
\label{tab:eff_accuracy}
\end{table*}

\begin{table}
\renewcommand\arraystretch{1.2}
    \centering
   
    \subfloat[Prompt update.\label{tab:abl-tpu}]{
    \begin{tabular}{p{0.32\linewidth}c}
        
        \toprule
        \bf \scriptsize Mode & acc. \\\midrule
        w/momentum & \bf 85.3 \\
        \\
        w/o momentum & 84.5 \\
        \bottomrule
    \end{tabular}
    }
    \hspace{+2pt}
    \subfloat[Comm. frequency\label{tab:abl-com}]{
    \begin{tabular}{p{0.26\linewidth}c}
        \toprule
        \bf \scriptsize epochs/round & acc. \\\midrule
        0.5 & \bf85.3 \\
        1 & \bf 85.3 \\
        2 & 84.6 \\\bottomrule
    \end{tabular}
    } 
     \caption{Ablation studies on different momentum update and communication frequency. }

\vspace{-0.5cm}
\end{table}
\textbf{Ablation on the different prompts structure.}
We classify prompts into 4 types (fixed, learnable, domain-specific, generative), and three representative SOTA models from baseline methods for comparison: PromptFL \cite{guo2023promptfl}, and ADAPT \cite{wei2024learning}, FedTPG \cite{qiu2024federated}. PromptFL represents methods that design a generator on the server side to produce prompts. ADAPT represents methods emphasizing domain-specific information. FedTPG represents methods focusing on designing single trainable networks to optimize prompts.

To validate the effectiveness of our approach and the improvements, several ablation models with different prompt structures were trained. These models include:
Our model with hand-crafted domain prompts (Ours-HDP), which replaces DSPs with structured prompts (i.e., "a photo of a [class]").
Our model with conditional soft prompts (Ours-CSP), which replaces DSPs with conditional soft prompts (i.e., $ t_i = [v]_1 [v]_2 \dots [v]_{M_1}$).
Our model without a generative model (Ours-WGM), which omits the second training step.
The results are presented in Table \ref{abla_prompt1}. Ours-HDP performs the worst, proving learnable prompt designing are crucial for performance. Although both ADAPT and OURS-WGM have domain-specific prompt, the improvement in results is not significant. However, Ours-CSP outperforms Ours-WGM and other baseline methods, underscoring the crucial role of the generative model in enhancing generalizability. Ultimately, our method achieves superior performance compared to all baselines, demonstrating the effective integration of DSPs and the generative model.

\textbf{Momentum update and communication frequency.} As elaborated in Section \ref{aggregation}, we employ a momentum-based approach to mitigate the performance degradation caused by large gradient variations. We conduct comparative experiments on parameters with and without momentum updates. As presented in Table \ref{tab:abl-tpu}, the results reveal a 0.8 percentage point decrease in accuracy without momentum. The experimental results for the momentum hyperparameter \( \alpha \) ranging from 0 to 1 are shown in Table \ref{tab:eff_accuracy}, with \( \alpha = 0.2 \) yielding the best performance. Additionally, we also test communication frequency, varying it to 0.5, 1, and 2 training epochs per communication round. Results (Table \ref{tab:abl-com}) showed no performance gain from more frequent aggregation compared to the default 1 epoch per round, while lower frequencies led to accuracy declines. This suggests that FedDSPG can achieve good performance by maintaining the default update frequency.

\begin{figure}
    \centering
    \includegraphics[width=1.0\linewidth]{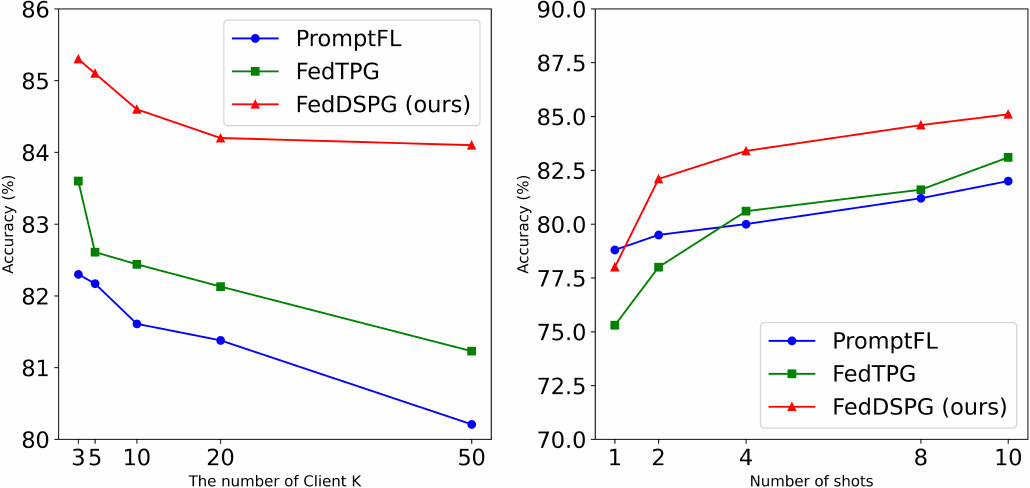}
    \caption{Ablation studies on different number of clients and shots.}
    \label{fig: result of num}
    \vspace{-0.4cm}
\end{figure} 
   
\textbf{Number of clients and shots.}
We conducted ablation studies on the number of clients and shots, the results are displayed in Figure \ref{fig: result of num}. Regarding the number of clients, although an inevitable performance drop occurs for all methods as the number of clients increases, FedDSPG exhibits only a marginal decline (approximately 1\%). For the number of shots, FedDSPG outperforms baseline methods when the number of shots exceeds one.

%% file: sec/6_conclusion.tex
\section{Conclusion}
This work introduces FedDSPG, a novel federated domain generalization approach that leverages domain-specific soft prompts (DSPs) to enhance model generalization in non-IID federated learning scenarios. By integrating domain and content knowledge into a generative model, FedDSPG generates diverse prompts for unseen target domains, improving performance in unknown domains. Through extensive evaluations, we demonstrate our method's superior performance against existing baselines, while also revealing the untapped potential of generative techniques in FDG scenarios. These findings provide both immediate practical value and long-term research directions for advancing FL in real-world applications.

%% file: main.bbl
\begin{thebibliography}{40}
\providecommand{\natexlab}[1]{#1}
\providecommand{\url}[1]{\texttt{#1}}
\expandafter\ifx\csname urlstyle\endcsname\relax
  \providecommand{\doi}[1]{doi: #1}\else
  \providecommand{\doi}{doi: \begingroup \urlstyle{rm}\Url}\fi

\bibitem[Bai et~al.(2024)Bai, Zhang, Li, Guo, Guo, Hou, Han, and Lu]{Bai2024DiPrompTDP}
Sikai Bai, Jiewei Zhang, Shuaicheng Li, Song Guo, Jingcai Guo, Jun Hou, Tao Han, and Xiaocheng Lu.
\newblock Diprompt: Disentangled prompt tuning for multiple latent domain generalization in federated learning.
\newblock \emph{Proceedings of the IEEE/CVF Conference on Computer Vision and Pattern Recognition}, 2024.

\bibitem[Bai et~al.(2025)Bai, Zhang, Zhou, Luan, and Chen]{bai2025soft}
Shuanghao Bai, Yuedi Zhang, Wanqi Zhou, Zhirong Luan, and Badong Chen.
\newblock Soft prompt generation for domain generalization.
\newblock In \emph{European Conference on Computer Vision}, pages 434--450. Springer, 2025.

\bibitem[Cui et~al.(2024)Cui, Li, Wang, and Shi]{cui2024harmonizing}
Tianyu Cui, Hongxia Li, Jingya Wang, and Ye Shi.
\newblock Harmonizing generalization and personalization in federated prompt learning.
\newblock \emph{International Conference on Machine Learning}, 2024.

\bibitem[Ding et~al.(2021)Ding, Wang, Xu, Welch, and Wang]{ding2021ccgan}
Xin Ding, Yongwei Wang, Zuheng Xu, William~J Welch, and Z~Jane Wang.
\newblock Ccgan: Continuous conditional generative adversarial networks for image generation.
\newblock In \emph{International Conference on Learning Representations}, 2021.

\bibitem[Dosovitskiy et~al.(2021)Dosovitskiy, Beyer, Kolesnikov, Weissenborn, Zhai, et~al.]{alexey2021image}
Alexey Dosovitskiy, Lucas Beyer, Alexander Kolesnikov, Dirk Weissenborn, Xiaohua Zhai, et~al.
\newblock An image is worth 16x16 words: Transformers for image recognition at scale.
\newblock In \emph{9th International Conference on Learning Representations, ICLR}, 2021.

\bibitem[Fang et~al.(2024)Fang, Yap, Lin, Zhu, and Liu]{fang2024source}
Yuqi Fang, Pew-Thian Yap, Weili Lin, Hongtu Zhu, and Mingxia Liu.
\newblock Source-free unsupervised domain adaptation: A survey.
\newblock \emph{Neural Networks}, 2024.

\bibitem[Feng et~al.(2021)Feng, You, Chen, Zhang, Zhu, et~al.]{feng2021kd3a}
Haozhe Feng, Zhaoyang You, Minghao Chen, Tianye Zhang, Minfeng Zhu, et~al.
\newblock Kd3a: Unsupervised multi-source decentralized domain adaptation via knowledge distillation.
\newblock In \emph{ICML}, pages 3274--3283, 2021.

\bibitem[Ge et~al.(2023)Ge, Huang, Xie, Lai, Song, Li, and Huang]{ge2023domain}
Chunjiang Ge, Rui Huang, Mixue Xie, Zihang Lai, Shiji Song, Shuang Li, and Gao Huang.
\newblock Domain adaptation via prompt learning.
\newblock \emph{IEEE Transactions on Neural Networks and Learning Systems}, 2023.

\bibitem[Guo et~al.(2023)Guo, Guo, Wang, Tang, and Xu]{guo2023promptfl}
Tao Guo, Song Guo, Junxiao Wang, Xueyang Tang, and Wenchao Xu.
\newblock Promptfl: Let federated participants cooperatively learn prompts instead of models-federated learning in age of foundation model.
\newblock \emph{IEEE Transactions on Mobile Computing}, 2023.

\bibitem[Ho et~al.(2020)Ho, Jain, and Abbeel]{ho2020denoising}
Jonathan Ho, Ajay Jain, and Pieter Abbeel.
\newblock Denoising diffusion probabilistic models.
\newblock \emph{Advances in Neural Information Processing Systems}, 33:\penalty0 6840--6851, 2020.

\bibitem[Hu et~al.(2021)Hu, Shen, Wallis, Allen-Zhu, Li, Wang, Wang, and Chen]{hu2021lora}
Edward~J Hu, Yelong Shen, Phillip Wallis, Zeyuan Allen-Zhu, Yuanzhi Li, Shean Wang, Lu Wang, and Weizhu Chen.
\newblock Lora: Low-rank adaptation of large language models.
\newblock \emph{9th International Conference on Learning Representations, ICLR}, 2021.

\bibitem[Hu et~al.(2024)Hu, Zhang, Xia, Chen, Luo, Sun, Wang, Qiao, Zeng, Sun, et~al.]{hu2024reclip}
Xuefeng Hu, Ke Zhang, Lu Xia, Albert Chen, Jiajia Luo, Yuyin Sun, Ken Wang, Nan Qiao, Xiao Zeng, Min Sun, et~al.
\newblock Reclip: Refine contrastive language image pre-training with source free domain adaptation.
\newblock In \emph{Proceedings of the IEEE/CVF Winter Conference on Applications of Computer Vision}, pages 2994--3003, 2024.

\bibitem[Jia et~al.(2021)Jia, Yang, Xia, Chen, Parekh, Pham, Le, Sung, Li, and Duerig]{jia2021scaling}
Chao Jia, Yinfei Yang, Ye Xia, Yi-Ting Chen, Zarana Parekh, Hieu Pham, Quoc Le, Yun-Hsuan Sung, Zhen Li, and Tom Duerig.
\newblock Scaling up visual and vision-language representation learning with noisy text supervision.
\newblock In \emph{International Conference on Machine Learning}, pages 4904--4916. PMLR, 2021.

\bibitem[Jia et~al.(2022)Jia, Tang, Chen, Cardie, Belongie, Hariharan, and Lim]{jia2022visual}
Menglin Jia, Luming Tang, Bor-Chun Chen, Claire Cardie, Serge Belongie, Bharath Hariharan, and Ser-Nam Lim.
\newblock Visual prompt tuning.
\newblock In \emph{European Conference on Computer Vision}, pages 709--727. Springer, 2022.

\bibitem[Li et~al.(2017)Li, Yang, Song, and Hospedales]{li2017deeper}
Da Li, Yongxin Yang, Yi-Zhe Song, and Timothy~M Hospedales.
\newblock Deeper, broader and artier domain generalization.
\newblock In \emph{Proceedings of the IEEE International Conference on Computer Vision}, pages 5542--5550, 2017.

\bibitem[Liu et~al.(2023)Liu, Xi, Li, Xu, Bai, and Zhao]{liu2023co}
Xinhui Liu, Wei Xi, Wen Li, Dong Xu, Gairui Bai, and Jizhong Zhao.
\newblock Co-mda: Federated multi-source domain adaptation on black-box models.
\newblock \emph{IEEE Transactions on Circuits and Systems for Video Technology}, 2023.

\bibitem[Liu et~al.(2024)Liu, Chen, Zhou, Xu, Xi, Bai, Zhao, and Zhao]{liu2024ufda}
Xinhui Liu, Zhenghao Chen, Luping Zhou, Dong Xu, Wei Xi, Gairui Bai, Yihan Zhao, and Jizhong Zhao.
\newblock Ufda: Universal federated domain adaptation with practical assumptions.
\newblock In \emph{Proceedings of the AAAI Conference on Artificial Intelligence}, pages 14026--14034, 2024.

\bibitem[Lu et~al.(2023)Lu, Hu, Wang, and Xie]{lu2023fedclip}
Wang Lu, Xixu Hu, Jindong Wang, and Xing Xie.
\newblock Fedclip: Fast generalization and personalization for clip in federated learning.
\newblock In \emph{ICLR Workshop on Distributed and Private Machine Learning}, 2023.

\bibitem[McMahan et~al.(2017)McMahan, Moore, , et~al.]{mcmahan2017communication}
Brendan McMahan, Eider Moore, , et~al.
\newblock Communication-efficient learning of deep networks from decentralized data.
\newblock In \emph{Artificial Intelligence and Statistics}, pages 1273--1282. PMLR, 2017.

\bibitem[Mirza and Osindero(2014)]{mirza2014conditional}
Mehdi Mirza and Simon Osindero.
\newblock Conditional generative adversarial nets.
\newblock \emph{Advances in Neural Information Processing Systems}, 2014.

\bibitem[Mohri et~al.(2019)Mohri, Sivek, and Suresh]{mohri2019agnostic}
Mehryar Mohri, Gary Sivek, and Ananda~Theertha Suresh.
\newblock Agnostic federated learning.
\newblock In \emph{International Conference on Machine Learning}, pages 4615--4625. PMLR, 2019.

\bibitem[Nayak et~al.(2022)Nayak, Yu, and Bach]{nayak2022learning}
Nihal~V Nayak, Peilin Yu, and Stephen~H Bach.
\newblock Learning to compose soft prompts for compositional zero-shot learning.
\newblock \emph{International Conference on Learning Representations}, 2022.

\bibitem[Peng et~al.(2019{\natexlab{a}})Peng, Bai, Xia, Huang, Saenko, and Wang]{peng2019moment}
Xingchao Peng, Qinxun Bai, Xide Xia, Zijun Huang, Kate Saenko, and Bo Wang.
\newblock Moment matching for multi-source domain adaptation.
\newblock In \emph{Proceedings of the IEEE/CVF International Conference on Computer Vision}, pages 1406--1415, 2019{\natexlab{a}}.

\bibitem[Peng et~al.(2019{\natexlab{b}})Peng, Huang, Zhu, and Saenko]{peng2019federated}
Xingchao Peng, Zijun Huang, Yizhe Zhu, and Kate Saenko.
\newblock Federated adversarial domain adaptation.
\newblock \emph{International Conference on Learning Representations, ICLR}, 2019{\natexlab{b}}.

\bibitem[Qiu et~al.(2024{\natexlab{a}})Qiu, Li, Mummadi, Ganesh, Li, Peng, and Lin]{qiu2023text}
Chen Qiu, Xingyu Li, Chaithanya~Kumar Mummadi, Madan~Ravi Ganesh, Zhenzhen Li, Lu Peng, and Wan-Yi Lin.
\newblock Text-driven prompt generation for vision-language models in federated learning.
\newblock \emph{International Conference on Learning Representations}, 2024{\natexlab{a}}.

\bibitem[Qiu et~al.(2024{\natexlab{b}})Qiu, Li, Mummadi, Ganesh, Li, Peng, and Lin]{qiu2024federated}
Chen Qiu, Xingyu Li, Chaithanya~Kumar Mummadi, Madan~Ravi Ganesh, Zhenzhen Li, Lu Peng, and Wan-Yi Lin.
\newblock Federated text-driven prompt generation for vision-language models.
\newblock In \emph{The Twelfth International Conference on Learning Representations}, 2024{\natexlab{b}}.

\bibitem[Radford et~al.(2021)Radford, Kim, Hallacy, Ramesh, Goh, Agarwal, Sastry, Askell, Mishkin, Clark, et~al.]{radford2021learning}
Alec Radford, Jong~Wook Kim, Chris Hallacy, Aditya Ramesh, Gabriel Goh, Sandhini Agarwal, Girish Sastry, Amanda Askell, Pamela Mishkin, Jack Clark, et~al.
\newblock Learning transferable visual models from natural language supervision.
\newblock In \emph{International Conference on Machine Learning}, pages 8748--8763. PMLR, 2021.

\bibitem[Shi et~al.(2024)Shi, Lu, Fang, and Zhang]{Shi2024UnsupervisedDA}
Kuo Shi, Jie Lu, Zhen Fang, and Guangquan Zhang.
\newblock Unsupervised domain adaptation enhanced by fuzzy prompt learning.
\newblock \emph{IEEE Transactions on Fuzzy Systems}, 32:\penalty0 4038--4048, 2024.

\bibitem[Song et~al.(2020)Song, Ma, Zhang, and Zhang]{song2020privacy}
Lei Song, Chunguang Ma, Guoyin Zhang, and Yun Zhang.
\newblock Privacy-preserving unsupervised domain adaptation in federated setting.
\newblock \emph{IEEE Access}, 8:\penalty0 143233--143240, 2020.

\bibitem[Tan et~al.(2024)Tan, Chen, Zhuang, Dong, Lyu, and Long]{tan2024heterogeneity}
Yue Tan, Chen Chen, Weiming Zhuang, Xin Dong, Lingjuan Lyu, and Guodong Long.
\newblock Is heterogeneity notorious? taming heterogeneity to handle test-time shift in federated learning.
\newblock \emph{Advances in Neural Information Processing Systems}, 36, 2024.

\bibitem[Venkateswara et~al.(2017)Venkateswara, Eusebio, Chakraborty, and Panchanathan]{venkateswara2017deep}
Hemanth Venkateswara, Jose Eusebio, Shayok Chakraborty, and Sethuraman Panchanathan.
\newblock Deep hashing network for unsupervised domain adaptation.
\newblock In \emph{Proceedings of the IEEE Conference on Computer Vision and Pattern Recognition}, pages 5018--5027, 2017.

\bibitem[Vettoruzzo et~al.(2024)Vettoruzzo, Bouguelia, Vanschoren, R{\"o}gnvaldsson, and Santosh]{alsulaimawi2024meta}
Anna Vettoruzzo, Mohamed-Rafik Bouguelia, Joaquin Vanschoren, Thorsteinn R{\"o}gnvaldsson, and KC Santosh.
\newblock Advances and challenges in meta-learning: A technical review.
\newblock \emph{IEEE Transactions on Pattern Analysis and Machine Intelligence}, pages 4763--4779, 2024.

\bibitem[Wei et~al.(2024)Wei, Wang, Shah, and Chellappa]{wei2024learning}
Guoyizhe Wei, Feng Wang, Anshul Shah, and Rama Chellappa.
\newblock Learning to prompt your domain for vision-language models.
\newblock In \emph{arXiv}, 2024.

\bibitem[Yang et~al.(2023)Yang, Wang, and Wang]{yang2023efficient}
Fu-En Yang, Chien-Yi Wang, and Yu-Chiang~Frank Wang.
\newblock Efficient model personalization in federated learning via client-specific prompt generation.
\newblock In \emph{Proceedings of the IEEE/CVF International Conference on Computer Vision}, pages 19159--19168, 2023.

\bibitem[Zaken et~al.(2021)Zaken, Ravfogel, and Goldberg]{zaken2021bitfit}
Elad~Ben Zaken, Shauli Ravfogel, and Yoav Goldberg.
\newblock Bitfit: Simple parameter-efficient fine-tuning for transformer-based masked language-models.
\newblock \emph{Annual Meeting of the Association for Computational Linguistics}, 2021.

\bibitem[Zhang et~al.(2023)Zhang, Gu, Matsuo, and Iwasawa]{zhang2023domain}
Xin Zhang, Shixiang~Shane Gu, Yutaka Matsuo, and Yusuke Iwasawa.
\newblock Domain prompt learning for efficiently adapting clip to unseen domains.
\newblock \emph{Transactions of the Japanese Society for Artificial Intelligence}, 38\penalty0 (6):\penalty0 B--MC2\_1, 2023.

\bibitem[Zhao et~al.(2023{\natexlab{a}})Zhao, Du, Li, Li, and Liu]{zhao2023fedprompt}
Haodong Zhao, Wei Du, Fangqi Li, Peixuan Li, and Gongshen Liu.
\newblock Fedprompt: Communication-efficient and privacy-preserving prompt tuning in federated learning.
\newblock In \emph{IEEE International Conference on Acoustics, Speech and Signal Processing (ICASSP)}, pages 1--5. IEEE, 2023{\natexlab{a}}.

\bibitem[Zhao et~al.(2023{\natexlab{b}})Zhao, Hu, Shao, and Hu]{zhao2023federated}
Ke Zhao, Junchen Hu, Haidong Shao, and Jiabei Hu.
\newblock Federated multi-source domain adversarial adaptation framework for machinery fault diagnosis with data privacy.
\newblock \emph{Reliability Engineering \& System Safety}, 236:\penalty0 109246, 2023{\natexlab{b}}.

\bibitem[Zhou et~al.(2022{\natexlab{a}})Zhou, Yang, Loy, and Liu]{zhou2022conditional}
Kaiyang Zhou, Jingkang Yang, Chen~Change Loy, and Ziwei Liu.
\newblock Conditional prompt learning for vision-language models.
\newblock In \emph{Proceedings of the IEEE/CVF Conference on Computer Vision and Pattern Recognition}, pages 16816--16825, 2022{\natexlab{a}}.

\bibitem[Zhou et~al.(2022{\natexlab{b}})Zhou, Yang, Loy, and Liu]{zhou2022learning}
Kaiyang Zhou, Jingkang Yang, Chen~Change Loy, and Ziwei Liu.
\newblock Learning to prompt for vision-language models.
\newblock \emph{International Journal of Computer Vision}, 130\penalty0 (9):\penalty0 2337--2348, 2022{\natexlab{b}}.

\end{thebibliography}
